\documentclass[runningheads]{llncs}
\usepackage{graphicx}
\usepackage{amsmath,amssymb} 
\usepackage{color}
\usepackage{algorithm}
\usepackage{algpseudocode}
\usepackage{graphics}
\usepackage{epsfig}
\begin{document}
\pagestyle{headings}
\mainmatter

\def\ACCV18SubNumber{196}  

\title{Generating Self-Guided Dense Annotations for Weakly Supervised Semantic Segmentation} 

\author{Zi-Yi Ke and Chiou-Ting Hsu}
\institute{Institute of Information Systems and Applications\\
National Tsing Hua University\\
Hsinchu, Taiwan}

\maketitle

\begin{abstract}
Learning semantic segmentation models under image-level supervision is far more challenging than under fully supervised setting. Without knowing the exact pixel-label correspondence, most weakly-supervised methods rely on external models to infer pseudo pixel-level labels for training semantic segmentation models. In this paper, we aim to develop a single neural network without resorting to any external models. We propose a novel self-guided strategy to fully utilize features learned across multiple levels to progressively generate the dense pseudo labels. First, we use high-level features as class-specific localization maps to roughly locate the classes. Next, we propose an affinity-guided method to encourage each localization map to be consistent with their intermediate level features. Third, we adopt the training image itself as guidance and propose a self-guided refinement to further transfer the image's inherent structure into the maps. Finally, we derive pseudo pixel-level labels from these localization maps and use the pseudo labels as ground truth to train the semantic segmentation model. Our proposed self-guided strategy is a unified framework, which is built on a single network and alternatively updates the feature representation and refines localization maps during the training procedure. Experimental results on PASCAL VOC 2012 segmentation benchmark demonstrate that our method outperforms other weakly-supervised methods under the same setting.
\end{abstract}

\section{Introduction}


Semantic image segmentation aims to densely assign a semantic label to every single pixel of an image. Although recent approaches \cite{R27}, \cite{R28}, \cite{R29}, \cite{R38}, \cite{R39}, \cite{R40}, \cite{R41} accelerate the development with convolutional neural networks (CNNs), their success was mostly attributed to numerous features learned from large-scale and densely-annotated training data. With dense (i.e., pixel-level) labels, existing fully supervised methods have learned to establish strong correspondences between semantic meaning and visual features. In addition, when further incorporating object detectors, many methods, such as \cite{R32}, \cite{R33}, \cite{R34}, \cite{R35}, take advantage of object-level knowledge and are able to identify and segment each individual object instance. However, as new semantic labels as well as new data are generated increasingly, it is by no means an easy task to obtain the pixel-level labels for new training data. To relieve the burden of obtaining dense labels, a number of weakly supervised methods have been proposed to train semantic segmentation models under weaker supervision, using either image-level labels \cite{R8}, \cite{R10}, \cite{R11}, \cite{R12}, \cite{R14}, \cite{R15}, \cite{R16}, \cite{R17}, \cite{R18}, \cite{R1}, \cite{R19}, \cite{R20}, \cite{R21}, \cite{R31}, \cite{R37}, \cite{R42}, \cite{R43}, \cite{R44}, \cite{R30}, points \cite{R7}, scribbles \cite{R6} or bounding boxes \cite{R4}, \cite{R5}.

\begin{figure*}[t]
\begin{minipage}[t]{1.0\linewidth}
  \centering
  \centerline{\includegraphics[height=5cm]{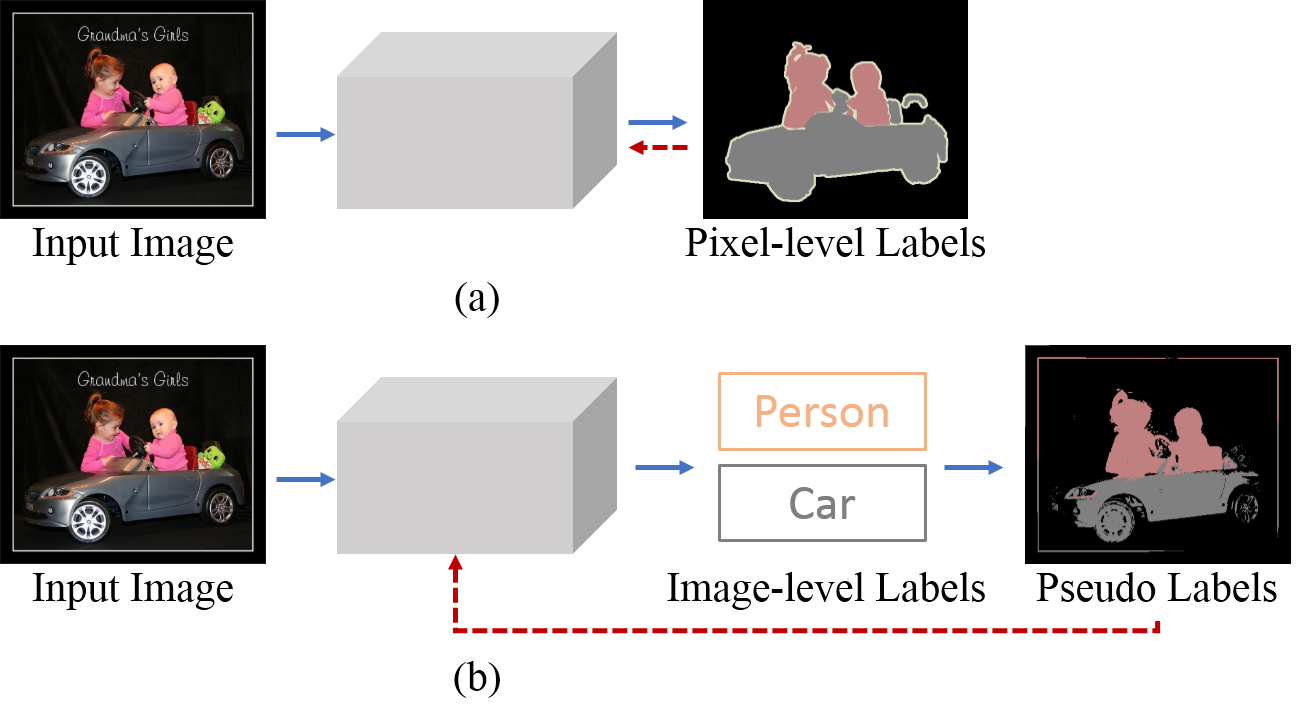}}
\end{minipage}
\caption{Illustrations of differences between training semantic segmentation models under (a) fully supervision with pixel-level labels, and (b) weak supervision with image-level labels.}

\label{fig:supervision}
\end{figure*}

Among the above-mentioned weak supervisions, image-level labels are the easiest one to collect but they provide the least supervision for training semantic segmentation models. As shown in Fig. \ref{fig:supervision} (a), in fully supervised learning, the pixel-level labels explicitly identify all the pixels belonging to each semantic class label; thus a well-designed neural network can learn to accurately locate and segment each individual class label. On the other hand, in Fig. \ref{fig:supervision} (b), in the image-level supervised learning, we are given only the semantic labels (i.e., person and car) but have no idea about the locations or shapes of these classes. When given no association between a class label and its corresponding set of pixels, it is very challenging to learn discriminative features and to obtain complete and precise segmentation results. Therefore, most existing methods \cite{R8}, \cite{R10}, \cite{R11}, \cite{R12}, \cite{R16}, \cite{R17}, \cite{R18}, \cite{R1}, \cite{R31}, \cite{R37}, \cite{R42}, \cite{R43}, \cite{R44}, \cite{R30} refer to external models to establish better pixel-label association. For example, visual saliency \cite{R11}, \cite{R12}, \cite{R18}, \cite{R1}, \cite{R31}, \cite{R42}, object proposals \cite{R10}, \cite{R11} and boundary recovery \cite{R17}, \cite{R8}, \cite{R44} have been included to infer pixel-level labels, which are then used as pseudo labels to learn a semantic segmentation network under fully supervision. However, these external models are usually trained under even stronger supervision (e.g., labels for training visual saliency or object proposal generators) or require additional computation to locate each class. Moreover, performance of these methods highly depends on the quality of information provided by the external models.

\begin{figure*}[t]
\begin{minipage}[t]{1.0\linewidth}
  \centering
  \centerline{\includegraphics[width=\textwidth]{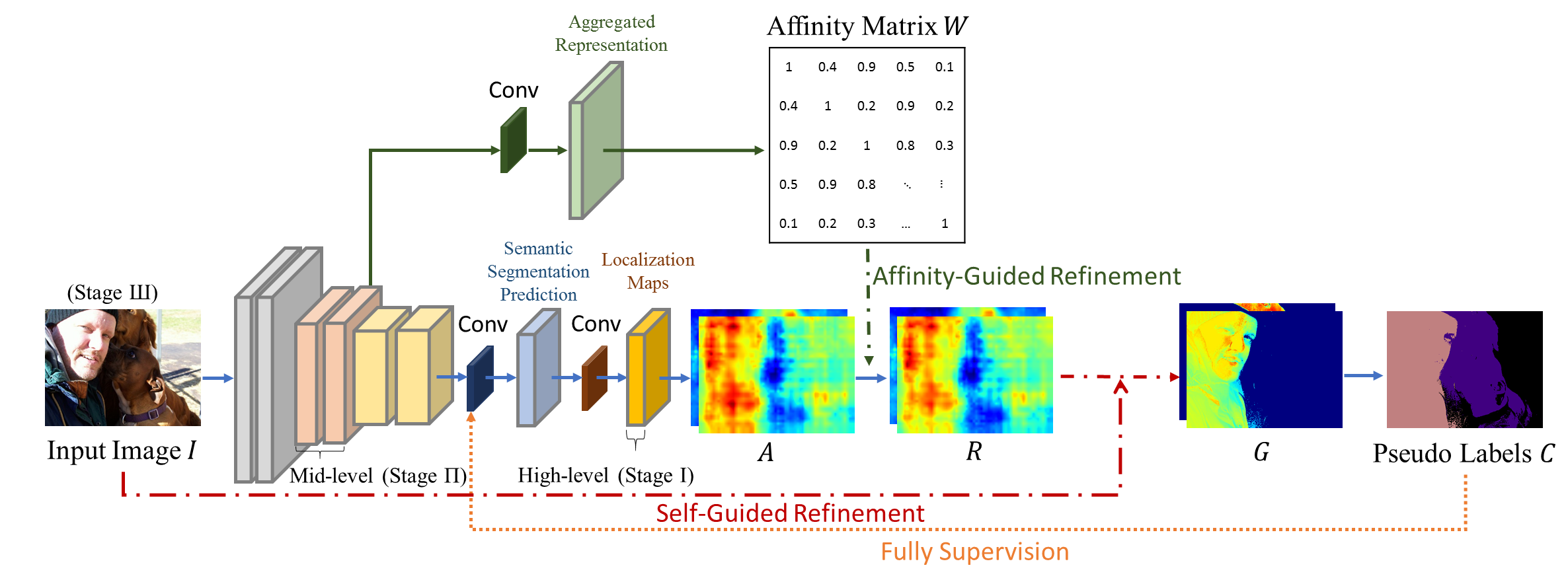}}
\end{minipage}
\caption{Generating pseudo pixel-level labels with the proposed self-guided strategy.}

\label{fig:flowchart}
\end{figure*}

\begin{figure*}[t]
\begin{minipage}[t]{1.0\linewidth}
  \centering
  \centerline{\includegraphics[width=\textwidth]{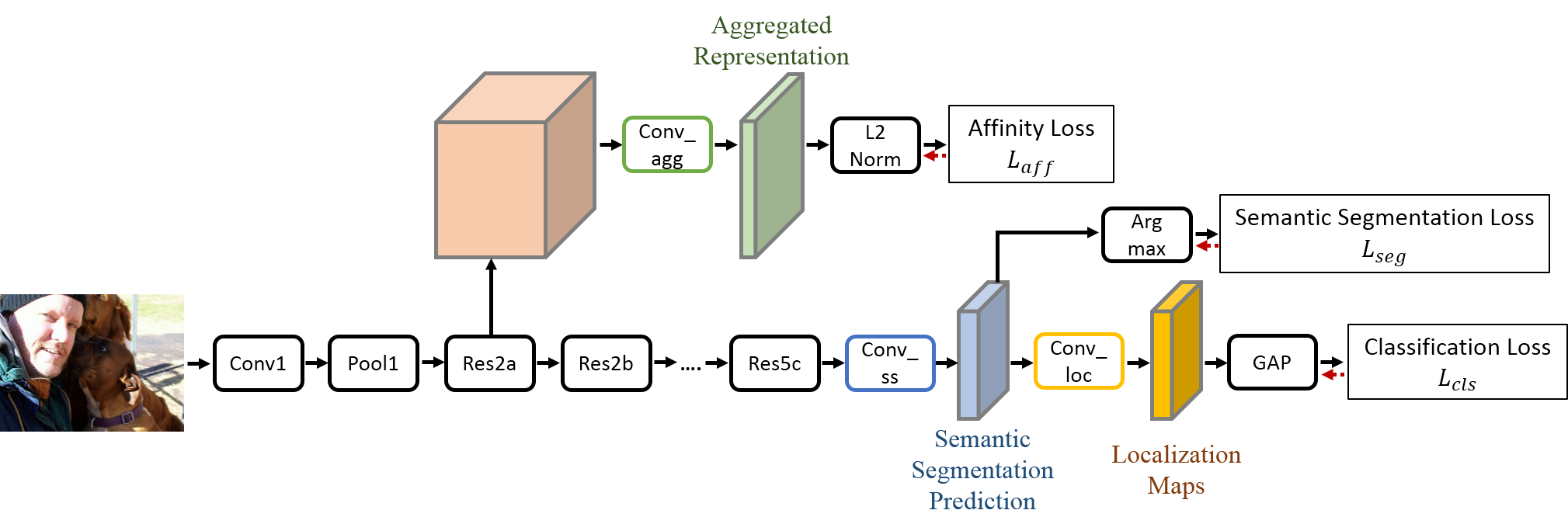}}
\end{minipage}
\caption{An illustration of the proposed training architecture.}

\label{fig:loss}
\end{figure*}

In this paper, our goal is to train a semantic segmentation network under image-level supervision without resorting to any other external models. We aim to fully benefit from the features learned during the training process and to jointly update the pixel-label association along with the feature learning. Our main idea comes from the observation that, an input image itself and its learned representation (even from weak labels) inherently preserve information across different scales and semantic levels. If we can comprehensively utilize this inherent information, there is no need to include any external or pre-trained models to infer the dense pseudo labels.    

Specifically, we propose to refer to the CNN layers backwardly so as to guide the generation of dense pseudo labels, by using the higher semantic layers to roughly locate each class label, using the intermediate representation layers to encourage consistency within each semantic label, and using the image itself to derive structure-preserving dense labels. This idea also mimic how we human beings learn to segment objects. For example, in Fig. \ref{fig:supervision} (b), when we are given the labels {\it person} and {\it car}, we tend to, by gathering information from all the training images containing these labels, first identify where the classes {\it person} and {\it car} locate. Then, we gradually learn to describe and finally learn to segment each individual class. Fig. \ref{fig:flowchart} illustrates the proposed idea. Because higher CNN layers better reflect image semantics, higher activation responses in these layers inherently indicate discriminative regions of the corresponding class label. Therefore, in the first stage, we use the activation responses in higher layers to predict the possibility of each pixel's belonging to a specific class. We call these pixel-wise soft maps as class-specific localization maps. Secondly, we rely on the intermediate feature representation learned in middle CNN layers to refine the localization maps. Although pixels belonging to the same semantic label may not always be visually similar, they tend to be consistent in the intermediate representation. We propose to include a convolution layer to learn feature affinities and then use the affinities to encourage feature-consistent activations within each localization map. Note that, we jointly learn the feature affinities along with image classification without referring to external supervision or models. Thirdly, to preserve detailed structure of each label, we use the image itself to guide the refining process. Inspired by the guided filtering \cite{R3}, we treat the localization maps as inputs and the image itself as the guide to derive final localization maps through structure transferring filtering. Finally, we determine pseudo pixel-level labels from the refined localization maps and then use the pseudo labels to train the semantic segmentation network under fully supervision. The proposed training architecture is shown in Fig. \ref{fig:loss}, where the network is alternatively trained using (1) image classification loss and affinity loss to generate pseudo labels, and (2) semantic segmentation loss to update the feature representation. Detailed training procedure will be described in Sec.4.

Our contributions are summarized as follows:

\begin{itemize}


\item Without resorting to any external models, we propose an unified framework with one single neural network to train the semantic segmentation model using only the image-level labels.


\item We propose a novel self-guided strategy to substantially utilize all the available information learned from the image itself to locate and refine the unknown pixel-level labels. 


\item We train the network alternatively between the classification and segmentation tasks, and also jointly learn the feature affinities along with the classification task. With this training procedure, we can enforce the learning of feature representation, feature affinities, and class-specific localization maps to mutually benefit from each other. 

\item Experimental results on PASCAL VOC 2012 segmentation benchmark show that our method outperforms existing methods under the same weakly supervised setting.


\end{itemize}


\section{Related Work}


Most image-level supervised methods, such as \cite{R19}, \cite{R20}, and \cite{R21}, rely on prior knowledge provided by pre-trained classification weights to determine pixel-label correspondence. However, using only the image class information, these methods tend to roughly locate each class and thus result in poor segmentation results. Therefore, many methods refer to external models to establish better pixel-label association. For example, the methods \cite{R11}, \cite{R12}, \cite{R1}, \cite{R18}, \cite{R31}, \cite{R42} adopt saliency detection to determine whether a pixel belongs to the foreground objects or not, and include such priors to supervise the learning. In \cite{R10} and \cite{R11}, a number of object proposals are detected, classified, and aggregated; then, the proposals with higher classification scores are treated as pseudo labels to learn the semantic segmentation. In \cite{R8}, an external CRF refinement network is jointly trained with semantic segmentation network so as to obtain edge preserving segmentation results. Other methods, such as \cite{R16}, \cite{R18}, \cite{R37}, \cite{R42}, \cite{R43}, \cite{R44}, \cite{R17}, extract local discriminative parts in the image classification task and then use these localization cues to establish better pixel-label correspondence for each foreground class.


However, inclusion of external models not only requires additional computation, the quality of pseudo labels also heavily depends on these models. Therefore, in \cite{R14}, the authors proposed to train a network for multi-label classification and use the derivatives of class scores in the intermediate convolutional layer as class saliency maps to estimate class-specific segmentation results. In \cite{R15}, the methods adopted feature maps extracted from early CNN layers to generate foreground/background map and then used the map to supervise the training. Nevertheless, because the derived pixel-label association is not precise enough, these methods can hardly learn discriminative features for each class and still have unsatisfactory segmentation results.

Among the related methods, \cite{R37} seems most related to ours, because we both propose to refine the pixel-label correspondence by referring to pixel-level affinity. In \cite{R37}, the authors proposed to train an AffinityNet and then to use AffinityNet to propagate the class activation responses. However, the framework in \cite{R37} included three deep neural networks and conducted complex procedures (e.g., sampling of pixel-pairs from confident areas, dealing with class-imbalance issue) to generate reliable labels for training AffinityNet. In contrast to \cite{R37}, we develop a single network to iteratively refine pseudo labels and conduct semantic segmentation. Moreover, we use only the original input image itself as ground truth to learn the feature affinities, and jointly learn the feature affinities along with the other learning tasks. In addition, we further include the input image itself as self-guidance to refine the missing detail after the affinity-guided refinement.
\section{Proposed Method}
\label{sec:blind}

We propose a novel self-guided strategy to generate pseudo labels for training semantic segmentation model in three stages. In the first stage, we extract coarse feature maps, which are called localization maps, in higher CNN layers to initialize rough pixel-label correspondence. In the second stage, we aggregate multiple feature maps in middle CNN layers to determine an affinity matrix, which encodes pixel-to-pixel similarities across different semantic levels, and then refine the localization maps under the guidance of this affinity matrix. In the third stage, we use the training image itself as guidance and further refine the localization maps by transferring the inherent structure from the guidance into the maps. Finally, we generate pseudo labels from the localization maps and train the semantic segmentation network under fully-supervised setting.


\subsection{Class-Specific Localization Maps}
\label{ssec:subhead1}


Let \(\mathcal{L}\) denote the label set with \(|\mathcal{L}-1|\) foreground object classes and one background class. We first build a general CNN with \(|\mathcal{L}|\) output maps \(A^c\), \(\forall{c\in\mathcal{L}}\) as the backbone semantic segmentation network. To jointly train the network as an image classifier, we add an additional convolutional layer (with \(|\mathcal{L}-1|\) filters of kernel size \(1\times1\times|\mathcal{L}|\)) followed by a global average pooling layer to predict the \(|\mathcal{L}-1|\) classification scores. Note that, because our goal is to locate the \(|\mathcal{L}-1|\) foreground classes, we do not employ a filter for the background class. Let \(E^c\) be the classification scores obtained by global average pooling and sigmoid function on \(A^c\). When training the CNN for image classification, we adopt the following loss function:


\begin{equation}
L_{cls}=-1(c\in{Z})\log(E^c)-1(c\notin{Z})\log(1-E^c), \forall{c\in\mathcal{L}\backslash {\it bg}}
\end{equation}
where {\it bg} is the background class, and \(Z\subseteq \mathcal{L}\backslash {\it bg}\) is the set of foreground classes present in the training image.

We call the \(|\mathcal{L}-1|\) output maps \(A^c\) as the class-specific localization maps and use these maps as our first guide to roughly locate each class. For example, in Fig. \ref{fig:fig2} (a) and (b), given an image \(I\) and its three class labels {\it bicycle}, {\it car} and {\it person}, we originally obtain the localization maps for all \(|\mathcal{L}-1|\) foreground classes. Since the three classes present in this image are given, we use only the maps for the \(|Z|=3\) present classes, i.e., \(A=\bigcup_{c\in{Z}}A^c\), in the following steps. In Fig. \ref{fig:fig2} (b), \(A^{c_1}\), \(A^{c_2}\) and \(A^{c_3}\) represent the localization maps of the classes {\it bicycle}, {\it car} and {\it person}, respectively, and each map roughly indicates the possibility of the locations' belonging to the corresponding class. 


\subsection{Affinity-Guided Refinement}
\label{ssec:aff}

Although higher CNN layers encode high-level semantic knowledge and indeed capture discriminative parts of each class, they only highlight class-specific locations with poor spatial resolution. As shown in Fig. \ref{fig:fig2} (a) and (b), even though the whole bicycle has consistent color, the initial localization map \(A^{c_1}\) of {\it bicycle} mostly highlights the central bicycle frame but is not equally activated in the wheels' region. As to the localization map \(A^{c_2}\) of {\it car}, the activated responses, though scatter around the cars, do not give a clear shape of cars.

To deal with the incomplete information in the initial localization maps \(A\), we propose to aggregate intermediate features and use the aggregated representation as guidance to refine the localization maps. Our idea is to encourage pixels of the same semantic label to have consistent responses in the corresponding class-specific localization map. The proposed affinity-guided refinement starts with the construction of an affinity matrix, which should reflect the semantic similarity of each pixel to every other pixels. As feature representation in middle CNN layers captures discriminative information across various scales and semantic levels, we propose to derive the affinity matrix in terms of the aggregated representation so as to enforce consistent responses for the same class.


To this end, we include a convolutional layer to aggregate the intermediate feature maps and jointly train this layer with the image classification task. We resize all the intermediate feature maps involved in this aggregation layer to \(w \times h\) and use a \(1 \times 1 \times k\) convolutional layer to aggregate these features across all channels. Let \(F\) denote the output feature maps of the aggregation convolutional layer, and \(W \in \mathbb{R}^{{wh} \times {wh}}\) be the affinity matrix defined by:
\begin{equation}
W_{pq}={\it exp}\{-\|F_p-F_q\|_2\} ,
\end{equation}
where \(W_{pq}\) is the feature affinity between pixels \(p\) and \(q\), and \(W\) is symmetric with \(W_{pp}=1\). Let \(T\) be the row-normalized affinity matrix expressed by \(T=D^{-1}W\), where \(D \in \mathbb{R}^{{wh} \times {wh}}\) is a diagonal matrix with \(D_{pp}=\sum_{q}W_{pq}\). Next, without adopting extra pixel-level supervision, we learn the aggregation layer by minimizing \(L_1\) distance between the normalized affinity matrix and the color similarity matrix directly measured from the input image itself. Let \(M \in \mathbb{R}^{{wh} \times {wh}}\) denote the color similarity matrix of the resized image \(I\), and each element \(M_{pq}\) encode the visual similarity between pixels \(p\) and \(q\) by:
\begin{equation}
M_{pq}=1-norm(\|I_p-I_q\|_2) ,
\end{equation}
where \(norm()\) is to normalize \(L_2\) distance to \([0,1]\). Then, we define the loss function for learning the aggregation layer by:
\begin{equation}
L_{aff}=\|T-M\|_1 .
\end{equation}

After obtaining the normalized affinity matrix \(T\), we adopt random walk process \cite{R36} to propagate local responses initialized in \(A\). The feature affinities in \(T\) indicate the transition probabilities on how the random walk modeling should propagate information. We thus diffuse the activated response through a single-step random walk, which is implemented via matrix multiplication by:
\begin{equation}
R=TA^{'} ,
\end{equation}
where \(A^{'} \in \mathbb{R}^{wh \times |Z|}\) is the reshaped \(A\) of size \({wh \times |Z|}\). From Equ. (5), the refined responses in the localization map \(R\) are weighted sum of initial responses from other locations according to the derived affinities.
As shown in Fig. \ref{fig:fig2} (c), the refined activations in  \(R\) better locate the bicycle wheels, and are much suppressed for the wrongly detected regions of {\it car}. As to the correctly located {\it person} class, the refined response remains intact. This example demonstrates that, through random walk process, we effectively enforce the localization maps to focus on compact and discriminative parts and also to suppress irrelevant or noisy responses.

\subsection{Self-Guided Refinement}
\label{ssec:fg}
\begin{figure}[t]
\begin{minipage}[t]{1.0\linewidth}
  \centering
  \centerline{\includegraphics[width=\textwidth]{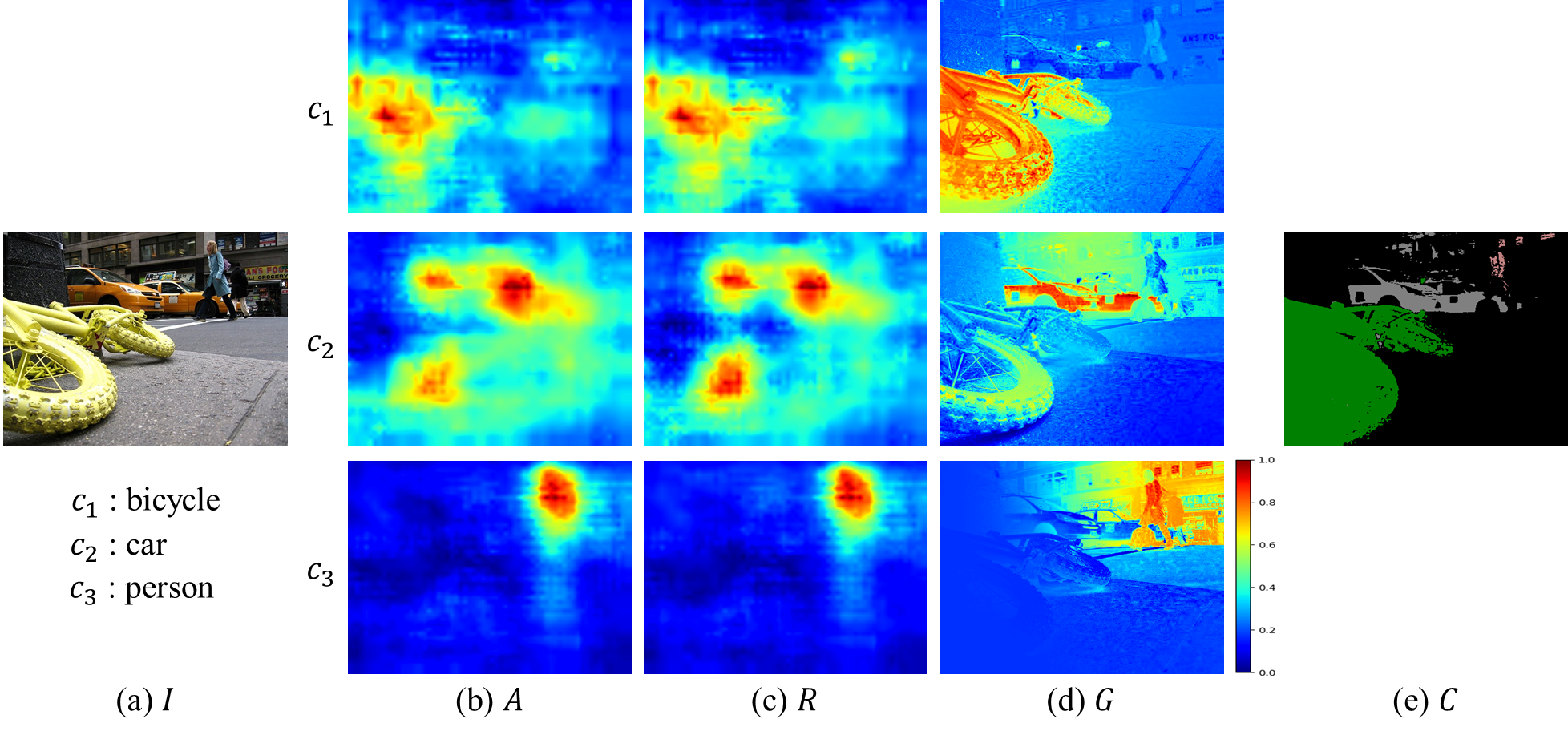}}
\end{minipage}
\caption{(a) An image labeled with classes {\it bicycle}, {\it car} and {\it person}, (b) the initial localization maps, (c) the localization maps after affinity-guided refinement, (d) the localization maps after affinity-guided and self-guided refinement, and (e) the pseudo pixel-level label.}
\label{fig:fig2}
\end{figure}

From Fig. \ref{fig:fig2} (c), the refined localization map \(R\) still gives no clear shape of each class and are far from ready to serve as dense class labels. We therefore need to resort to other available information to revise the localization maps. We are inspired by the guided filter \cite{R3}, which can faithfully transfer the structure of a guidance image into an input and can recover the object boundaries of an input through a fast and locally-linear operation. Therefore, we propose to use the training image itself as guidance and the localization maps as inputs to the guided-filtering operations.


We first reshape \(R\) into \(w \times h \times |Z|\) and conduct Otsu's thresholding on \(R^c\), \(\forall{c\in Z}\) to obtain its binary maps \(B^c\), where a pixel \(p\) with \(B^c_{p}\)=1 is assumed to belong to the class \(c\). Next, we use \(B^c\) as the input and the training image \(I\) as the guide, and conduct the guided filtering for each class \(c\) independently to refine the incomplete localization in \(B^c\). In \cite{R3}, the filtered output \(G^c\) is assumed to be a linear model of \(I\) within a local window:
\begin{equation}
G^c_{p}={\alpha}_{p}I_{p}+{\beta}_{p} , \forall_{p} ,
\end{equation}
where \({\alpha}_{p}\) and \({\beta}_{p}\) are the guided filter coefficients for the pixel \(p\). By minimizing the difference between \(G^c\) and \(B^c\), the authors in \cite{R3} explicitly derived the coefficients \({\alpha}_{p}\), \({\beta}_{p}\) and the filtered output \(G^c\) at the pixel \(p\) by:
\begin{equation}
G^c_{p}=\sum_{q}K_{pq}(I)B^c_{q} ,
\end{equation}
where
\begin{equation}
K_{pq}(I)=\frac{1}{|\omega|^2}\sum_{j:(p,q)\in \omega_{j}}(1+\frac{(I_p-\mu_j)(I_q-\mu_j)}{\sigma^2_j+\epsilon}), 
\end{equation}
\(q\) denotes all the other pixels located within the local window \(\omega_j\) around \(p\), \(\mu_j\) and \(\sigma^2_j\) are the mean and variance of \(I\) in \(\omega_j\), and \(\epsilon\) is a regularization parameter. More derivation details can be found in \cite{R3}.

Since the guided filtering output \(G^c_{p}\) is a linear transform of \(I_{p}\), we can implement the guided filtering by a trainable convolutional layer. With this trainable layer, we can simultaneously update the coefficients \({\alpha}_{p}\) and \({\beta}_{p}\) along with the refined \(R\). 
When back propagating the loss residual to the previous layer, we update the derivatives with respect to \(G^c\) and compute the derivatives with respect to \(\alpha\) and \(\beta\) by

\begin{equation}
\frac{\partial L_{seg}}{\partial \alpha_{p}} \gets {I_p}\frac{\partial L_{seg}}{\partial G^c_{p}}+\frac{\partial L_{seg}}{\partial \alpha_{p}}, 
\end{equation}
\begin{equation}
\frac{\partial L_{seg}}{\partial \beta_{p}} \gets \frac{\partial L_{seg}}{\partial G^c_{p}}+\frac{\partial L_{seg}}{\partial \beta_{p}},
\end{equation}
where \(L_{seg}\) denotes the semantic segmentation loss function and will be defined in the next subsection.

When leveraging the detailed structure in the guidance image, as shown in Fig. \ref{fig:fig2} (d), we greatly preserve object boundaries in the localization maps \(G\).
%
\subsection{Pseudo Pixel-Level Label}
\label{ssec:mask}
Next, we simply assign the pixel-level label as the class \(c\) with the strongest response in \( G^c\) for each pixel. We initialize all the pixel-level label with the background label, and then determine the pseudo label at pixel \(p\) by:
\begin{equation}
C_{p}=\underset{c\in\{x|G^x_{p}>\tau\} }{\arg\max}G^c_{p} ,
\end{equation}
where \(\tau\) is the foreground/background threshold determined by Otsu's method.

Using the pseudo pixel-level label \(C\), we then train the network under fully-supervised semantic segmentation setting with the single-label cross-entropy function:
\begin{equation}
L_{seg}=-\sum_{p\in \mathcal{P}}\log(S^{C_{p}}_p) ,
\end{equation}
where \(\mathcal{P}\) is the set of pixels of an image and \(S^{C_{p}}_p\) is the softmax probability of the class \(C_p\).

\subsection{Training Strategy}
\label{ssec:trainingstratagy}


The overall training procedure conducts two steps alternatively so as to minimize different loss functions. In step \(1\), we train the network by optimizing the function:
\begin{equation}
\min \sum_{I \in \mathcal{I}} L_{cls}(I)+\lambda L_{aff}(I) ,
\end{equation}
to jointly learn class-specific localization maps and the affinity matrix. Next, we conduct affinity-guided refinement and self-guided refinement on the localization maps and derive pseudo pixel-level labels from these maps. In step \(2\), we minimize the semantic segmentation loss by using the pseudo labels as ground truth:
\begin{equation}
\min \sum_{I \in \mathcal{I}}L_{seg}(I) .
\end{equation}
Detailed training procedure is summarized in Algorithm 1. 

\section{Experiments}
\subsection{Experimental Settings}

\label{ssec:expsetup}
{\bf Dataset and Evaluation Metrics} We evaluate our approach on PASCAL VOC 2012 segmentation dataset \cite{R25}, which consists of 20 foreground classes and one background class. The dataset contains 1464 training images, 1449 validation images and 1456 test images. We follow the common practice to augment the training set to 10582 images with additional data \cite{R45}. During the training phase, we utilize only the image-level labels. The segmentation performance is measured by the mean intersection-over-union (mIoU) between the ground truth and the predicted segmentation over the 21 classes. To fairly compare with other methods, we also provide segmentation results with CRF post-processing \cite{R26}.
\linebreak {\bf Implementation} We adopt DeepLab-V2 \cite{R22}, whose parameters are initialized by ResNet-101 \cite{R23} pretrained on ImageNet \cite{R24}, as our network. Then we use 'poly' learning policy and randomly crop images into \(321 \times 321\) to train the network. As described in Algorithm 1, we update the network by alternatively optimizing different objectives. When training the image classification task coupled with feature affinity learning, we initialize the three convolution layers (which we additionally include for segmentation prediction, localization maps prediction, and feature aggregation) to be Gaussians of zero mean, \(0.01\) standard deviation and bias with zeros. We use the intermediate level features {\it res2a-relu} as the input to the aggregation convolutional layer to encode the \(256\) features into \(k=3\) channels. The batch size and initial learning rate are set as 10 and 0.001, respectively; and the initial class-specific localization maps for each image are obtained after \(50k\) iterations. When training the semantic segmentation task, we conduct the training for \(10k\) iterations, and set the batch size and initial learning rate as 8 and 0.0005, respectively. 
\linebreak {\bf Parameter Setting} We set \(n=2\) and \(\lambda=1\) in Algorithm 1. In the affinity-guided refinement, we resize the feature maps into \(w=50\) and \(h=50\) for computational efficiency. When implementing the guided filtering, we set the filtering window size and regularization parameter \(\epsilon\) in Equ. (8) as \(35\) and \(10^{-6}\), respectively.

\subsection{Evaluation of the Proposed Method}
\label{ssec:method}
To justify the effectiveness of different components in our approach, we evaluate the performance of semantic segmentation when training using different pseudo labels as ground truth. In Table 1, we use \(A\), \(R\), (\(G\)-w/o aff), \(G\), (\(G\)-iter15-w/o aff), and (\(G\)-iter15) to denote the pseudo labels obtained from the initial localization map \(A\), from the affinity-guided refined map \(R\), from the map obtained by guided filtering on \(A\), from the map \(G\), from the map obtained by 15 iterations of self-guided refinement on \(A\), and from the map obtained by affinity-guided refinement followed by 15 iterations of self-guided filtering, respectively.

{\bf Effectiveness of each component} As shown in Table 1, when including affinity-guided refinement, we improve the segmentation performance from \(42.7\%\) (\(A\)) to \(43.5\%\) (\(R\)). This improvement verifies that, by encouraging consistent activation responses within each class, we indeed improve the quality of pseudo labels even without knowing the object boundaries. When conducting the self-guided refinement on \(R\), we originally have degraded performance from \(43.5\%\) (\(R\)) to \(43.1\%\) (\(G\)). We suspect that, if some foreground shapes in \(R\) are too noisy and become dissimilar with those in the original image, even the structure-preserving filter can hardly rectify the situation in one single step. Especially, once surrounding pixels of certain foreground classes in \(R\) are wrongly preserved in \(G\), the network trained by \(G\) would result in inferior performance than trained by \(R\). Nevertheless, after conducting multiple times of self-guided refinement, we can better transfer the object structure from the guidance to the localization maps. Our experiments show that, after about 15 iterations, the refined localization maps (\(G\)-iter15) correctly capture the object structure with clearer boundaries and shapes, and the performance of \(G\)-iter15 is increased by \(4\%\) to \(47.3\%\).

\begin{figure}[t]
\begin{minipage}[t]{1.0\linewidth}
  \centerline{\includegraphics[width=\textwidth]{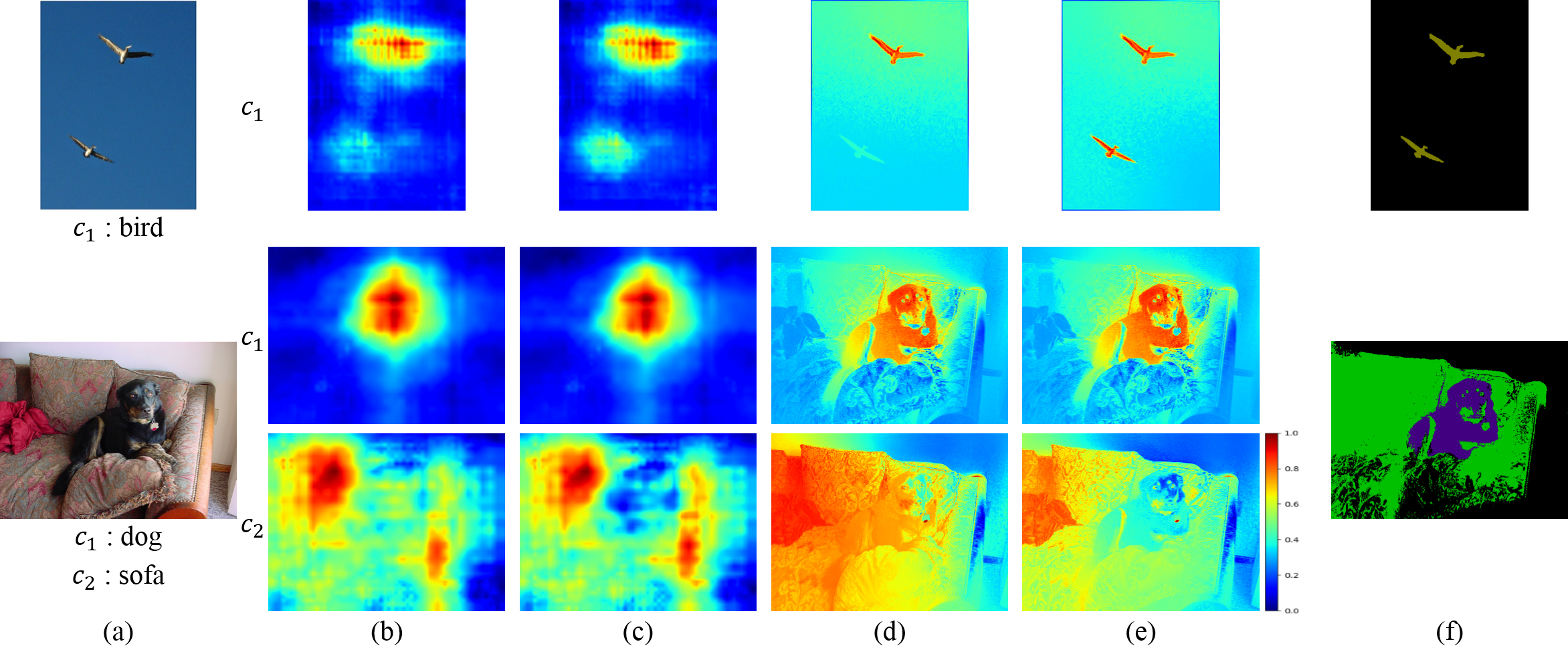}}
\end{minipage}
\caption{(a) The input image \(I\), (b) the initial localization maps \(A\), (c) the localization maps after affinity-guided refinement \(R\), (d) the localization maps after self-guided refinement (\(G\)-iter15-w/o aff), and (e) the localization maps after affinity-guided refinement and self-guided refinement (\(G\)-iter15), and (f) the pseudo label.}
\label{fig:fig4}

\end{figure}

{\bf Evaluation of affinity-guided refinement} In Fig. \ref{fig:fig4}, we show two examples to demonstrate the effectiveness of affinity-guided refinement. As shown in Fig. \ref{fig:fig4} (a) and Fig. \ref{fig:fig4} (b), although there are two birds in the image, the bottom one has much smaller activation responses than the top one in \(A\). In the second example, the discriminative parts of the {\it sofa} class do not form a compact cluster but are occluded by the {\it dog}. If, without affinity-guided refinement, as shown in Fig. \ref{fig:fig4} (d), the self-guided refinement completely smooths out the bottom {\it bird} in the first example, and results in highly overlapped maps for {\it sofa} and {\it dog} in the second example. On the other hand, by adopting the proposed affinity-guided refinement, we propagate active responses using random walk process and successfully compensate the missing regions in \(A\). As shown in Fig. \ref{fig:fig4} (c), the bottom {\it bird} has stronger activations in \(R\), and the occluded region of {\it sofa} has reduced response whereas the correctly highlighted {\it dog} remains intact. After the self-guided refinement, as shown in Fig. \ref{fig:fig4} (e), we further reshape the discriminative parts of each class in \(R\) by referring to the original images and obtain good structure-preserved responses. Therefore, as shown in Fig. \ref{fig:fig4} (f), the pseudo labels determined from \(G\)-iter15 establish reliable pixel-label correspondence and achieve improved segmentation performance.

\begin{table}[t]
\begin{minipage}[t]{0.5\linewidth}
\label{tab:table1}
\caption{Evaluation of the proposed method on PASCAL VOC 2012 validation set.}
  \centering
\begin{tabular}{ c|c|c }
\multicolumn{1}{c|}{}& \multicolumn{2}{c}{mIOU \((\%)\)}\\
 \hline
 pseudo labels
 & w/o CRF
 & w/ CRF\\
 \hline
 \(A\)
 & 42.7 & 45.3 \\  
 \(R\)
 & 43.5 & 46.3 \\  
 \(G\)-w/o aff
 & 41.1 & 43.5 \\
 \(G\)
 & 43.1 & 45.8 \\
 \(G\)-iter15-w/o aff
 & 46.8 & 49.4 \\
 \(G\)-iter15
 & 47.3 & 50.5 \\

\end{tabular}
\end{minipage}
\begin{minipage}[t]{0.5\linewidth}
  \begin{algorithmic}
    \State \underline{{\bf Algorithm 1} Proposed iterative procedure}
    \For{$i=1$ to $n$}
       \State {\bf Step \(1\):}
       \State $//$ Generate pseudo labels
            \State \(\min \sum_{I \in \mathcal{I}}L_{cls}(I)+\lambda L_{aff}(I)\)
            \State Initialize localization maps;
            \State Affinity-guided refinement;
            \State Self-guided refinement;
            \State Determine pseudo labels;
       \State {\bf Step \(2\):}
       \State $//$ Train for semantic segmentation
            \State \(\min \sum_{I \in \mathcal{I}}L_{seg}(I)\)
    \EndFor
    \label{code:recentEnd}
  \end{algorithmic}
 \end{minipage}
\end{table}
\subsection{Comparisons with Existing Methods}
\label{ssec:img-level}
In Table \ref{tab:table2}, we compare the proposed method with existing image-level supervised approaches on PASCAL VOC 2012 validation set \cite{R25}. The results in Table \ref{tab:table2} show that, these methods \cite{R1}, \cite{R31}, \cite{R18}, \cite{R12} significantly benefit from extra saliency detectors to derive accurate pseudo pixel-level labels. However, as mentioned before, the saliency detectors need to be additionally trained upon bounding boxes or pixel-level class-agnostic labels. Thus, these methods implicitly rely on additional labels, which may involve even stronger supervision than image-level labels, to achieve the excellent performance. Similarly, the methods \cite{R10}, \cite{R11}, which assembled MCG object proposals as pseudo labels, are not only time-consuming for producing large amounts of proposals; their comparison with other methods seems fundamentally unfair, because the MCG generator is trained under full supervision on PASCAL VOC. The method in \cite{R8} benefited from a CRF-RNN boundary refinement network, which is jointly trained with the semantic segmentation network to infer boundary information, and improve the performance by \(3\%\). Some other methods \cite{R12}, \cite{R30}, \cite{R31} collected additional training data to improve the performance, but these data involve strong supervision as well. In \cite{R37}, though not including external data or model, the authors developed a rather complex framework to learn semantic affinity from selected pairs of data, and then to combine information from three separate models. Unlike \cite{R37}, we propose a simple network and comprehensively utilize the learned features with different representative information as guidance to generate pseudo labels.

As to other methods which also include no external training data or models, the methods \cite{R19}, \cite{R20} and \cite{R21}, by relying on pre-trained image classifier weights to derive rough pixel-label correspondence, can hardly achieve satisfactory performance. In \cite{R14}, although the model is also trained for image classification task and directly employed for semantic segmentation, the method did not include further learning to locate each class. Therefore, there is a \(4\%\) performance gap between \cite{R14} and our method. This performance gap also verifies that, an accurate pixel-label correspondence is indeed critical to weakly-supervised learning. The method in \cite{R15}, which learns pixel-label mask from the network's built-in feature maps, can only differentiate the foreground from background and thus leads to inferior performance.

\begin{figure}[t]
\begin{minipage}[t]{1.0\linewidth}
  \centering
  \centerline{\includegraphics[width=\textwidth]{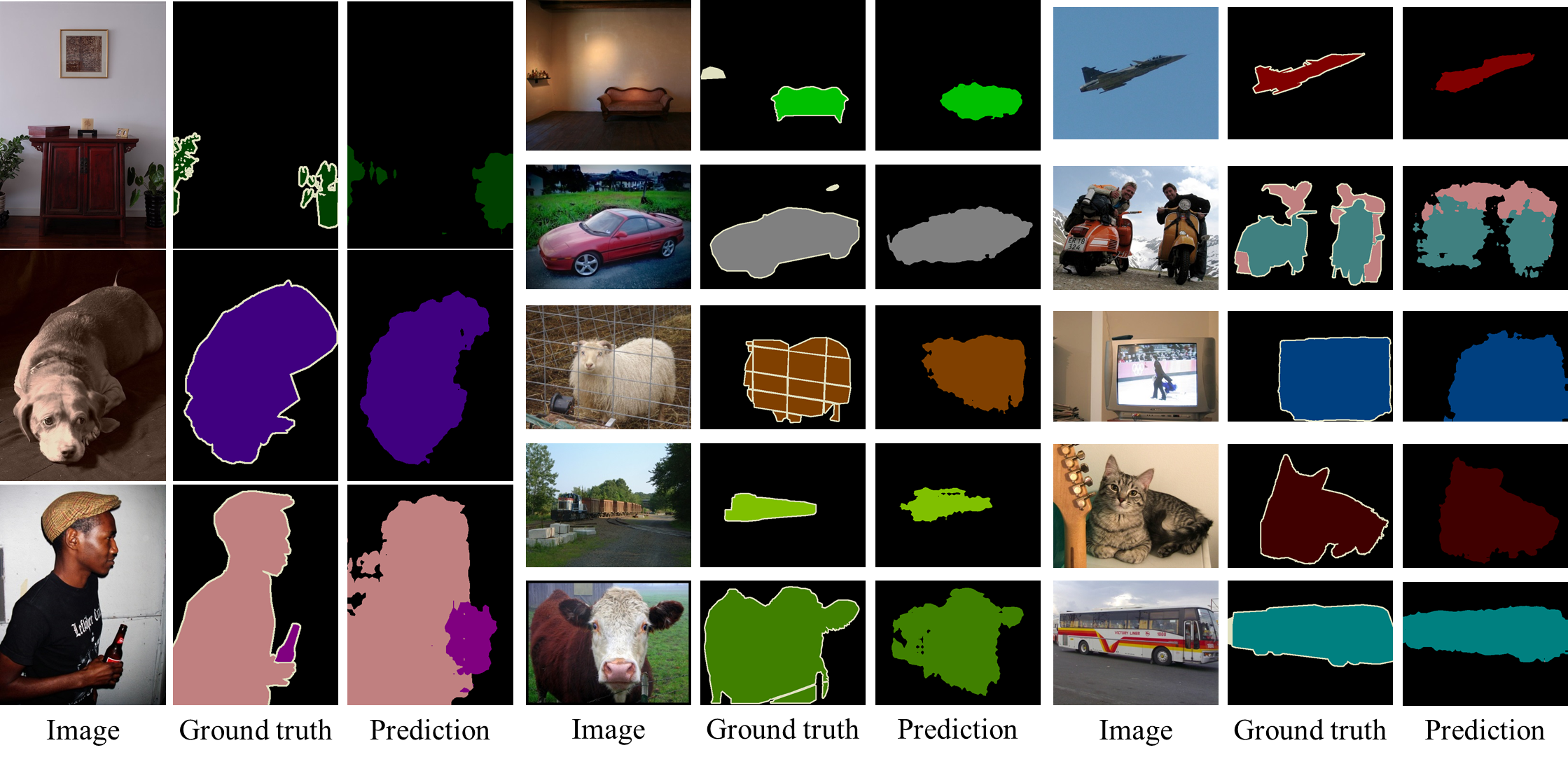}}
\end{minipage}
\caption{Segmentation results on PASCAL VOC 2012 validation set.}
\label{fig:exp-fig}

\end{figure}

Fig. \ref{fig:exp-fig} shows qualitative segmentation results and demonstrates the effectiveness of the proposed method for most classes. More per class performance is given in Table \ref{tab:table-class}.


\begin{table}[t]
\caption{Per class IOU on PASCAL VOC 2012 test set of the proposed method.}
\label{tab:table-class}
\begin{minipage}[h]{1.0\linewidth}
  \centering
  \resizebox{\textwidth}{!}{
\begin{tabular}{ ccccccccccccccccccccc|c }
 
 bg&aero&bike&bird&boat&bottle&bus&car&cat&chair&cow&table&dog&horse&mbike&person&plant&sheep&sofa&train&tv&mIOU \((\%)\)\\
 \hline
 78.6&46.5&19.8&62.7&20.0&47.2&61.8&53.0&67.3&16.7&49.4&36.5&57.1&50.9&63.2&52.5&37.2&51.8&40.0&42.8&36.1&47.2
 
\end{tabular}}
\end{minipage}
\end{table}
\begin{table}[htb]
\caption{Comparisons with existing methods on PASCAL VOC 2012 validation set.}
\label{tab:table2}
\begin{minipage}[htb]{1.0\linewidth}
  \centering
  \begin{tabular}{ lcc|cc }
 \multicolumn{1}{l}{}&
 \multicolumn{1}{c}{}&
 \multicolumn{1}{c|}{}&
 \multicolumn{2}{c}{mIOU \((\%)\)}\\
 \hline
 Methods & Extra data & External model & w/o CRF  & w/ CRF\\
 \hline
 AffiniyNet (ResNet38) \cite{R37} & - & AffinityNet & - & 61.7 \\
 AffiniyNet (DeepLab) \cite{R37} & - & AffinityNet & - & 58.4 \\
 DCSP (ResNet101) \cite{R1} & - & Saliency & 59.5 & 60.8 \\
 DCSP (VGG16) \cite{R1} & - & Saliency & 56.5 & 58.6 \\
 SaliencyNet (DeepLab) \cite{R31} & \(10K Box\) & Saliency & 51.2 & 55.7 \\
 AE-PSL (DeepLab) \cite{R18} & - & Saliency & - & 55.0 \\
 AugFeed (DeepLab) \cite{R10} & - & MCG & 50.4 & 54.3 \\
 CombiningCues (DeepLab) \cite{R8} & - & CRF-RNN & - & 52.8 \\
 TranserNet \cite{R30} & \(60K Pixel\) & - & - & 52.1 \\
 SEC (DeepLab) \cite{R17} & - & - & 44.3 & 50.7 \\
 STC (DeepLab) \cite{R12} & \(40K Image\) & Saliency & - & 49.8 \\
 BFBP (VGG16) \cite{R15} & - & - & 44.8 & 46.6 \\
 DCSM (VGG16) \cite{R14} & - & - & 40.5 & 44.1 \\
 SN-B (DeepLab) \cite{R11} & - & MCG/Saliency & - & 41.9 \\
 WSSL-img (DeepLab) \cite{R19} & - & - & - & 38.2 \\
 MIL w/ILP \cite{R21} & - & - & 32.6 & - \\ 
 MIL (VGG16) \cite{R20} & - & - & 25.0 & - \\  
 \hline
 \(G\)-iter15 (DeepLab) & - & - & 46.2 & 48.7\\
 \(G\)-iter15 (ResNet-101) & - & - & {\bf 47.3} & {\bf 50.5} 
\end{tabular}
\end{minipage}
\end{table}

\section{Conclusion}
\label{sec:majhead}
In this paper, we propose to train a weakly-supervised semantic segmentation network through generating reliable pixel-level annotations. Without resorting to any external information, our proposed self-guided strategy refers to the training image itself as well as the learned feature representation to derive accurate pixel-label correspondence. The proposed framework is built upon a single neural network and is alternatively trained by optimizing different objectives. Our experimental results on PASCAL VOC 2012 segmentation benchmark outperform other methods under the same weakly supervised setting. Quantitative and qualitative results also verify the effectiveness of the proposed model.
\bibliographystyle{splncs}
\bibliography{egbib}

\end{document}